\documentclass[10pt,twocolumn,letterpaper]{article}

\usepackage{cvpr}
\usepackage{times}
\usepackage{epsfig}
\usepackage{graphicx}
\usepackage{amsmath}
\usepackage{amssymb}
\usepackage{subfigure}
\usepackage{multirow}
\usepackage{bigstrut}
\usepackage{color}
\definecolor{amber}{rgb}{1.0, 0.75, 0.0}
\usepackage{ifthen}
\definecolor{gray}{rgb}{0.5,0.5,0.5}
\definecolor{green}{rgb}{0, 0.4, 0}
\definecolor{orange}{rgb}{1, 0.4, 0}
\definecolor{mahogany}{rgb}{0.75, 0.25, 0.0}
\definecolor{purple}{rgb}{0.6, 0, 0.6}
\definecolor{darkgreen}{rgb}{0, 0.4, 0}
\definecolor{frenchblue}{rgb}{0.0, 0.45, 0.73}
\definecolor{crimson}{rgb}{0.6, 0.0, 0.0}
\definecolor{amber}{rgb}{1.0, 0.75, 0.0}
\newboolean{revising}
\setboolean{revising}{True}
\ifthenelse{\boolean{revising}}
{
	\newcommand{\ignore}[1]{}

}
{
	\newcommand{\ignore}[1]{}

}
\usepackage[breaklinks=true,bookmarks=false]{hyperref}

\cvprfinalcopy % *** Uncomment this line for the final submission

\def\cvprPaperID{7952} % *** Enter the CVPR Paper ID here

\begin{document}

%%%%%%%%% TITLE
\title{Explainable Object-induced Action Decision for Autonomous Vehicles}

\author{Yiran Xu \quad Xiaoyin Yang \quad Lihang Gong \quad Hsuan-Chu Lin \\ Tz-Ying Wu \quad Yunsheng Li \quad Nuno Vasconcelos \\
Department of Electrical and Computer Engineering\\
University of California, San Diego\\
{\tt\small \{y5xu,x4yang,lgong,lhsuanch,tzw001,yul554,nuno\}@ucsd.edu} 
% For a paper whose authors are all at the same institution,
% omit the following lines up until the closing ``}''.
% Additional authors and addresses can be added with ``\and'',
% just like the second author.
% To save space, use either the email address or home page, not both
}

\maketitle
% \thispagestyle{empty}

%%%%%%%%% ABSTRACT
\begin{abstract}
    A new paradigm is proposed for autonomous driving. The new paradigm lies between the end-to-end and pipelined approaches, and is inspired by how humans solve
    the problem. While it relies on scene understanding, the latter only considers objects that could originate hazard. These are denoted as action inducing, since changes in their state should trigger vehicle actions. They also define a set of explanations for these actions, which should be produced jointly with the latter. 
    An extension of the BDD100K dataset, annotated for a set of $4$ actions and $21$ explanations, is proposed. A new multi-task formulation of the problem, which optimizes the accuracy of both action commands and explanations, is then introduced. A CNN architecture is finally proposed to solve this problem, by combining reasoning about action inducing objects and global scene context. Experimental results show that the requirement of explanations improves the recognition of action-inducing objects, which in turn leads to better action predictions. 
\end{abstract}

%%%%%%%%% BODY TEXT
\section{Introduction}

Deep learning has enabled enormous progress
in autonomous driving. Two major approaches
have emerged. {\it End-to-end\/} systems~\cite{kim2017interpretable,xu2017end,kim2018textual,
  wang2019monocular,wang2019deep} map
the visual input directly into a driving action, such as ``slow-down''
or ``turn''. {\it Pipelined\/} systems first detect
objects and obstacles, and then use that information to decide on
driving actions. Both approaches have advantages and shortcomings.
End-to-end systems are theoretically optimal, since all
visual information is used for decision-making. By the data processing
theorem~\cite{cover2012elements}, 
 intermediate decisions can only lead to loss
of information and compromise end-to-end optimality. However,
end-to-end predictors are complex, requiring
large datasets and networks. 
Pipelines have the advantage of modularity,
decomposing the problem into a collection of much smaller sub-problems,
such as object detection, trajectory analysis and planning, etc.
This approach has spanned sub-literatures in 3D object
detection \cite{ku2019monocular, li2019stereo,
  meyer2019lasernet, wang2019pseudo}, segmentation \cite{li2019attention},
depth estimation \cite{liu2019neural, ranjan2019competitive}, 3D
reconstruction \cite{lin2019photometric}, among other topics. Nevertheless,
the failure of a single module can compromise the performance of the whole
system~\cite{crash2019enabling}.

\begin{figure}[t]
    \centering
    \includegraphics[scale=0.29,trim={0.5cm 0.5cm 0.5cm 0.5cm},clip]{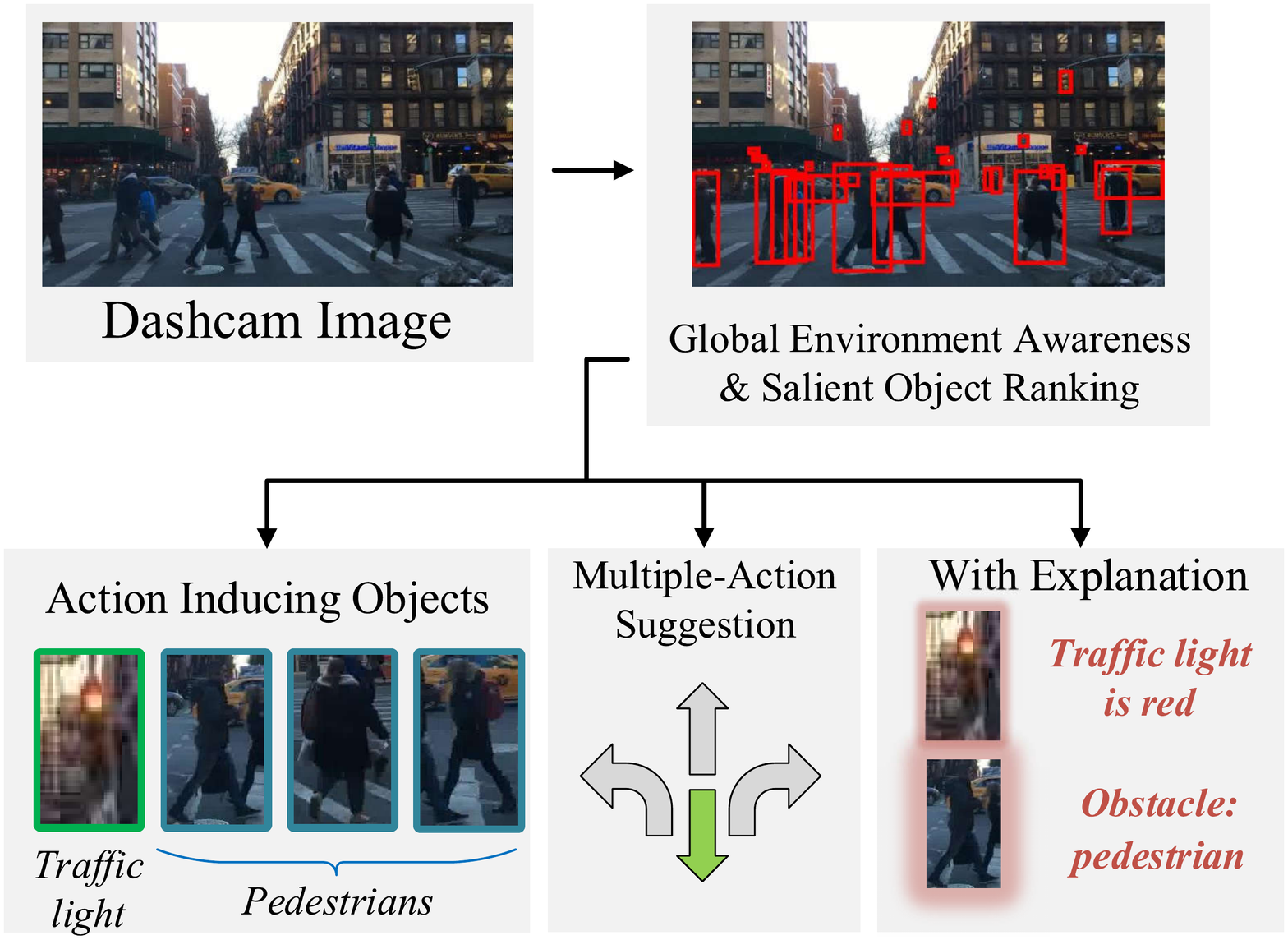}
    \caption{Top: while autonomous vehicles face complex scenes, composed of
      many objects, only a few of these are action-inducing.
      Bottom: each action-inducing object has an associated explanation
      for the related action. The  arrows represent actions ``move forward'', ``turn left'', ``stop/slow down'', and ``turn right'' (count-clockwise order).  {\color{green} Green} identifies the acceptable action.
      %In this example, the action of ``slow down''
      %can be explained by wither ``a red light'' or ``a pedestrian
      %that crosses the street''.}
      }
    \label{fig:intro}
    \vspace{-1.5em}
\end{figure}

Beyond driving performance, interpretability is a major concern for autonomous
driving. End-to-end systems are notoriously poor in this
regard given their black box nature. While they can be complemented
with explanations~\cite{vedantam2017context,anne2018grounding}, these are not yet fully reliable. 
Pipelined systems are more amenable
to forensic analysis, by analysing the performance of each
module and assessing its potential contribution to a system failure.
However, current pipelined approaches are not designed for this. Instead,
each module becomes a computer vision problem
of its own, sometimes with its own datasets and leaderboards. This makes
it easy to loose track of the overall goal when evaluating performance.
For example, further progress on the detection of occluded pedestrians on
sidewalks is unlikely to improve autonomous driving
performance.

In this work, we advocate for the design of systems in between these
two ends of the spectrum. This is
inspired by how humans solve the problem. When deciding
between slowing down or turning, humans do not
employ a strict end-to-end strategy. Instead, they perform a certain amount
of understanding and reasoning about scene objects. This, however,
is far from full blown scene understanding. Instead, they only
pay attention to objects directly related to the driving task.
For example, when driving, most people pay little attention to pedestrians
on the sidewalk, or parked cars, focusing
instead on objects that could create
hazard \cite{xia2018predicting, palazzi2018predicting}.
We denote these as {\it action-inducing objects\/} because changes in their
state, \eg a pedestrian that crosses the street, trigger vehicle actions, \eg ``slow down'' or ``stop''. This is illustrated in Figure~\ref{fig:intro}.
While the scene contains many objects, only a few are action-inducing.

One advantage of focusing on action-inducing objects
is that they define the set of explanations for driving actions.
For example, in Figure~\ref{fig:intro}, a car approaching an intersection
slows down due to two action-inducing objects:
a red traffic light and pedestrians crossing the street.
More generally, {\it every
action-inducing object has an associated
explanation for the action it induces.\/} This implies that only a finite
universe of explanations is required, and that explanations can be viewed as
an auxiliary set of semantic classes, to be predicted simultaneously
with the actions. This leads naturally to a multi-task problem, where
the goal is to jointly predict actions and explanations, as shown in
Figure~\ref{fig:intro}.

In this work, we propose  an architecture for this joint prediction.
We start by introducing the BDD object induced action (BDD-OIA) dataset,
for research in this problem.
A limitation of current driving datasets
\cite{geiger2012we,caesar2019nuscenes,Cordts2016Cityscapes,
  neuhold2017mapillary} is that they are too simple, \ie it is frequently
easy to determine the action to take. To address this problem, BDD-OIA
only includes complicated scenes where multiple actions are possible.
These are manually annotated for actions and associated explanations.
BDD-OIA is complemented by a new architecture for joint
action/explanation prediction, implemented by a multi-task
CNN that leverages the Faster R-CNN to detect objects and a global scene
context module to determine which of these are action-inducing, outputing
the associated pair of actions and explanations.

The multi-task formulation has several nice properties. First,
rather than an extra burden on classifier design, explanations become a
secondary source of supervision. By forcing the classifier to predict
the action ``slow down'' because ``the traffic light is red,''
the multi-task setting exposes the classifier to the causality
between the two. This is much richer
supervision than an image simply labelled as ``slow down''. Second,
unlike prediction heatmaps computed a
posteriori \cite{kim2017interpretable,kim2018textual,xia2018predicting, palazzi2018predicting,
  simonyan2013deep,zeiler2014visual,
  hong2015decoupled,selvaraju2017grad,huk2018multimodal}, or the
synthesis of full blown natural language~\cite{kim2018textual} with recurrent
networks or generative systems, a finite set of explanations can be mapped into
a classification task. Hence, the system can be explicitly optimized for
generation of natural language explanations while posing a relatively
simple learning problem.
In fact, our experiments show that, under the proposed multi-task formulation,
the generation of explanations {\it improves\/} the action prediction
performance of the network. To the best of our knowledge, this is the
first showing that explainable AI can improve the performance of a
autonomous driving system. The proposed network is also shown to achieve
good performance on the task or predicting driving commands and explanations, 
achieving state of the art results on BDD-OIA.
% outperforming previous approaches [IS THIS DONE?].

Overall, the paper makes four main contributions.
\begin{itemize}
  \vspace{-.08in}
\item A large dataset annotated for both driving commands and explanations.
    \vspace{-.08in}
  \item A new multi-task formulation of the action prediction problem
    that optimizes for both the accuracy of action commands and explanations.
    \vspace{-.08in}
    \item A CNN architecture for the solution of this problem, combining
      reasoning about action-inducing objects and global scene context.
      \vspace{-.08in}
    \item An experimental evaluation showing that the generation of
      explanations improves the decision making for actions, and both
      benefit from a combination of object-centric and global scene reasoning.
\end{itemize}
%-------------------------------------------------------------------------
\section{Related work} \label{relatedwork}

\paragraph{End-to-end learning for autonomous driving.} End-to-end driving was first
proposed in 1989, with the ALVINN system \cite{alvinn1989}. 
\cite{nvidia2016} later demonstrated the strong ability of CNNs to produce
steering wheel commands. These systems are strictly end-to-end, using
no explicit reasoning about objects. More recently a number of
approaches \cite{kim2017interpretable,xu2017end,kim2018textual,
  wang2019monocular,wang2019deep} to end-to-end learning for
vehicle control have taken advantage of both context and object features.
% Others \cite{chen2015,sauer2018} proposed affordance indicators to guide
% vehicles through complicated driving environments. 
However, many of these
systems are trained on driving simulators \cite{wang2019monocular,
  wang2019deep}. Despite progress in domain adaptation algorithms, systems
trained on synthetic data tend to underperform when deployed in the real
world. Our proposed network is an end-to-end system fully trained on real images, leveraging object
detection and contextual reasoning.
% which produces both commands and explanations.
\vspace{-1.5em}

\paragraph{Global-local contextual representations.}Contextual relationships between global and local scene
features are important cues for the identification of the important parts of a scene. 
% for a given task. 
Contextual reasoning has a long history in computer
vision \cite{alexe2012searching,gkioxari2015contextual,Devin2018deep,
  cai2016unified,yin2018zoom,yang2018graph,wang2018videos}. For example,
\cite{cai2016unified} shows that multi-scale pooling improves the
object detection performance of the Faster R-CNN\cite{ren2015faster} and 
\cite{yin2018zoom} relies on contextual information to recognize
visual relationships. 
However, contextual learning has received limited
attention in the autonomous driving literature. \cite{wang2019deep}
proposed a selector of the most critical objects in the scene, but neglected
the importance of global features altogether. \cite{kim2017interpretable,
  kim2018textual} considered instead features extracted from the whole scene,
ignoring objects. Our proposed architecture accounts for both objects and
context, exploring their relationships and detailed supervision, in the form of explanations, to separate action-inducing objects from objects unimportant for driving. 
\vspace{-1.5em}

\paragraph{Attention mechanisms.} Attention mechanisms have been widely utilized in neural networks~\cite{xu2015show,wang2017residual}. Attention maps are also used to visualize the inner workings of these networks  \cite{simonyan2013deep,
  zeiler2014visual,hong2015decoupled, selvaraju2017grad, huk2018multimodal}.
In the autonomous driving realm,
\cite{bojarski2016} developed a richer notion of attention on pixels that
collaboratively contribute to the prediction. Studies of human attention,
using eye tracking systems, have also been performed in
\cite{xia2018predicting, palazzi2018predicting} to determine which scene
regions capture the gaze of a driver. \cite{kim2017interpretable,
  kim2018textual} introduced the concept of visual attention maps for
end-to-end driving. Instead of pixel-level attention, \cite{wang2019deep} proposed an object-level attention model. This could be seen as a weaker form of the now proposed idea of using objects to define actions. 
\vspace{-1.5em}

\paragraph{Explanations.} Textual explanations are sometimes used for insight on the network understanding of images or scene \cite{vedantam2017context,anne2018grounding,lecun2015,xu2015show}. 
For example, \cite{xu2015show,vedantam2017context,anne2018grounding} generate text to explain either attention maps or network predictions. In the autonomous driving setting, Kim \etal \cite{kim2018textual} integrate textual generation and an attention mechanism with end-to-end driving. An attention-based video-to-text model is used to generate human understandable explanations for each predicted action. The  formulation now proposed, based on action-inducing objects,  enables one-hot encoded explanations. This eliminates the ambiguity of textual explanations, and improves action prediction performance.
\vspace{-1.5em}

\paragraph{Datasets.}
Several autonomous driving datasets, contain both real images or video
and information from multiple sensors, including radar, LiDAR, GPS or IMU
information. KITTI \cite{geiger2012we}
was one of the earliest to be annotated with object bounding boxes,
semantic segmentation labels, depth, and 3D point clouds.
BDD100K \cite{yu2018bdd100k} contains 100K videos annotated with
image level labels, object bounding boxes, drivable areas, lane markings,
and full-frame instance segmentation. Apolloscape
\cite{Huang_2018_CVPR_Workshops} has 140K images, RGB videos and
corresponding dense 3D point clouds with focus on 3D labeling and 
semantic segmentation. nuScenes \cite{caesar2019nuscenes} contains
1000 scenes with sensor information produced by camera,
LiDAR, and radar. While large and richely annotated, none of these datasets
addresses the detection of action-inducing objects. 
The  dataset now proposed is derived from BDD100K but contains substantial extra annotations to enable this objective.

\section{Joint Action and Explanation Prediction}

In this section, we introduce the problem of jointly predicting
and explaining object induced actions.

\subsection{Definitions} \label{task}

Smart driving systems perform two major classes of actions.
Actions in the first class are independent of
other objects in the environment. For example, a car navigating on a desert
freeway can simply decide to slow down to optimize energy consumption.
These actions do not require sophisticated perception and are not
hazardous. Actions in the second class involve
reasoning about other objects on the road or its surroundings, as
illustrated in Figure~\ref{fig:intro}. 
While we refer to them as
object induced actions, the definition of
object could be abstract. For example, a lane change
may be possible due to an open lane ``object.'' For the purposes of
this work, any object or event that can be detected or recognized
by a vision system is considered an object. 

One of the interesting properties of object induced actions is their
strong causal structure. For example, in Figure~\ref{fig:intro},
the pedestrians that cross the street force the car to slow down.
While there can be multiple causes for the action, \eg the traffic light
is also red, the cardinality of this set is relatively small. This
implies that the action has a small set of possible explanations.
If the car ran the intersection, it must have either not detected the
pedestrians or the traffic light. While ``corner'' cases
can always exist, \eg the car failed to detect a broken tree limb in the
middle of the road, these can be incrementally added to the set of
explanations. 
In any case, because the set of objects and
explanations is relatively small, the joint prediction of actions and
explanations can be mapped into a factorial classification problem. 

\begin{table}[t]
  \centering
    \scriptsize	
    % \resizebox{\columnwidth}
    \begin{tabular}{c|c|c|c}
    \hline
    \textbf{Action Category} & \textbf{Number} & \textbf{Explanations} & \textbf{Number} \bigstrut\\
    \hline
    \multirow{3}[2]{*}{Move forward} & \multirow{3}[2]{*}{12491} & Traffic light is green & 7805 \bigstrut[t]\\
          &       & Follow traffic & 3489 \\
          &       & Road is clear & 4838 \bigstrut[b]\\
    \hline
    \multirow{6}[2]{*}{Stop/Slow down} & \multirow{6}[2]{*}{10432} & Traffic light  & 5381 \bigstrut[t]\\
          &       & Traffic sign & 1539 \\
          &       & Obstacle: car & 233 \\
          &       & Obstacle: person & 163 \\
          &       & Obstacle: rider & 5255 \\
          &       & Obstacle: others & 455 \bigstrut[b]\\
    \hline
    \multirow{6}[4]{*}{Turn left} & \multirow{3}[2]{*}{838} & No lane on the left & 150 \bigstrut[t]\\
          &       & Obstacles on the left lane & 666 \\
          &       & Solid line on the left & 316 \bigstrut[b]\\
\cline{2-4}          & \multirow{3}[2]{*}{5064} & On the left-turn lane & 154 \bigstrut[t]\\
          &       & Traffic light allows & 885 \\
          &       & Front car turning left & 365 \bigstrut[b]\\
    \hline
    \multirow{6}[4]{*}{Turn right} & \multirow{3}[2]{*}{1071} & No lane on the right & 4503 \bigstrut[t]\\
          &       & Obstacles on the right lane & 4514 \\
          &       & Solid line on the right & 3660 \bigstrut[b]\\
\cline{2-4}          & \multirow{3}[2]{*}{5470} & On the right-turn lane & 6081 \bigstrut[t]\\
          &       & Traffic light allows & 4022 \\
          &       & Front car turning right & 2161 \bigstrut[b]\\
    \hline
    \end{tabular}%
     \vspace{0.6em}
      \caption{
      Action and explanation categories in the BDD-OIA dataset. Because actions are objectet induced, explanations are based on objects. Changing lanes to left/right is merged with turn left/right to avoid distribution imbalance. For the turn left/right rows, the upper sub-row presents statistics of changing lane to the left/right and the lower sub-row those of turning left/right. For these actions, explanations address why the action is not possible.
      }
      \vspace{-1em}
  \label{tab:all_labels}%
  
\end{table}%

In this work, we consider the set of 4 actions commonly predicted by
end-to-end driving systems~\cite{xu2017end,wang2019deep}, and listed in the left of
Table~\ref{tab:all_labels}. These are complemented by the 21 explanations
listed in the right side of the table. Different from previous works, we consider the classification of actions to be multi-label classification, \ie we can have more than one choice. Mathematically, given an image $I$ or
a video $V$ in some space $\cal X$, the goal is to determine the best
action $A \in \{0,1\}^4$ to take and the explanation $E \in \{0,1\}^{21}$ that
best justifies it. This is implemented by the mapping
\begin{equation}
  \phi: \mathcal{X} \mapsto (A,E) \in \{0,1\}^4 \times  \{0,1\}^{21}.
\end{equation}
For instance, if the possible actions are ``Stop'' and ``Change to the left
lane'', then $A = [0,1,1,0]^T$. The structure of the action and explanation
label vectors is defined in Table~\ref{tab:all_labels}.
In summary, joint action/explanation prediction is a combination
of two multi-label classification problems. 

\subsection{BDD-OIA Dataset}\label{dataset}

In the real world, driving is composed of long periods with very little to do (car simply ``moves forward'') and relatively short periods where the driver must decide between a set of object induced actions.
When decisions have to be made, they are more difficult if
environments are complex, \eg{} with road obstacles, pedestrian crossings, etc.
Yet, driving datasets contain a relatively low percentage of such
scenes. This can be seen from
Table~\ref{tab:density}, which summarizes the average densities of
pedestrians and moving vehicles per image of several datasets. The fact that these numbers are low suggests that most of the driving scenarios are relatively simple. Previous research also only predicts the action chosen by the driver \cite{xu2017end,wang2019monocular,wang2019deep}, creating the false impression that only that action was possible.
All of this, makes existing datasets poorly suited to study object induced
actions. Beyond this, because these datasets are not annotated with
explanations for object induced actions, they cannot be used to learn how
to generate such explanations. 

\begin{figure}[t]
    \centering
    \includegraphics[width=.95\linewidth]{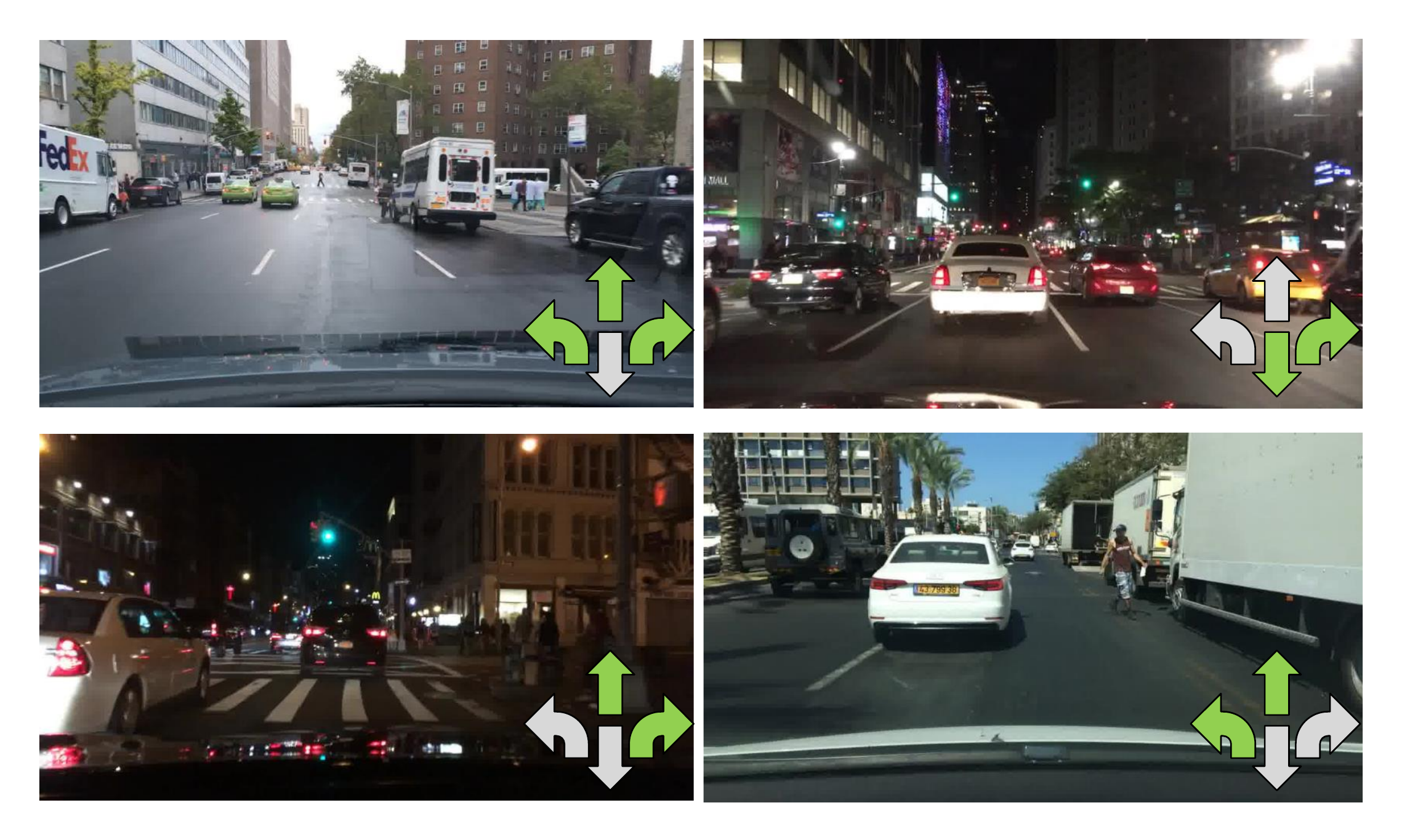}
    \caption{Scenes in BDD-OIA. The green arrows in the bottom right show the ground truth for possible actions.}
    \label{fig:annotations}
    \vspace{-1.1em}
\end{figure}

To address these problems, we selected a subset of
BDD100K~\cite{yu2018bdd100k} video clips containing at least 5 pedestrians
or bicycle riders and more than 5 vehicles. To increase scene diversity,
these videos were selected under various weather conditions and times of the
day.  This resulted in 22,924 5-second video clips, which were annotated on
MTurk for the 4 actions and 21 explanations of Table~\ref{tab:all_labels}. We refer to this dataset as the {\it BDD Object Induced
  Actions\/} (BDD-OIA) dataset. Figure \ref{fig:annotations} shows examples
of typical scenes in BDD-OIA. These are all complex driving scenes,
where multiple action choices are frequently possible. There are also
many objects, \eg cars parked on the side of the street, that are not action
inducing and a few object inducing objects per scene. action-inducing
objects can be other vehicles, pedestrians, traffic
lights, or open lanes.

\begin{table}[t]
  \begin{center}
    \small
    \begin{tabular}{c|c|c}
    \hline
    Dataset & \# pedestrians &  \# vehicles \\
    \hline
    BDD100K \cite{yu2018bdd100k} & 1.2 & 9.7 \\
    % ApolloScape \cite{Huang_2018_CVPR_Workshops} & 16.8 & 38.1  \\
    KITTI \cite{geiger2012we} & 0.8 & 4.1  \\
    Cityscapes \cite{Cordts2016Cityscapes} & 7.0 & 11.8 \\
    % Mapillary Vistas \cite{neuhold2017mapillary} & ~2.8 & ~6.0 \\
    BDD-OIA  & \textbf{8.0} & \textbf{11.8}\\
    \hline
    \end{tabular}
\end{center}
\vspace{-0.2em}
\caption{Densities of pedestrians and vehicles per image in popular driving datasets (statistics based on training set). On average, the scenes of the proposed BDD-OIA dataset are more complicated than those of previous datasets.}
\label{tab:density}
\vspace{-1em}
\end{table}

Table~\ref{tab:all_labels} summarizes
the dataset content in terms of
the 4 action classes and the 21 explanation categories. The coverage of actions is fairly balanced. In fact, our initial goal was to include the four BDD classes
(``move forward,'' ``stop/slow,'' ``left,'' and ``right turn,'') plus the two
novel classes of ``change lane'' to the left/right. However, there are very few opportunities to turn in BDD100K. To avoid a highly unbalanced dataset,
we merged turns and lane changes.
The coverage of the 21 explanation categories is a lot more unbalanced.
The most probable is ``Traffic light is green,'' (7805 occurrences),
while the rarest are ``No lane on the left'' (150) and ``On the left-turn lane'' (154). 
\vspace{-0.5em}

\begin{figure*}[htbp]
\begin{center}
\includegraphics[scale=0.55,trim={0.5cm 7cm 0.2cm 7.2cm},clip]{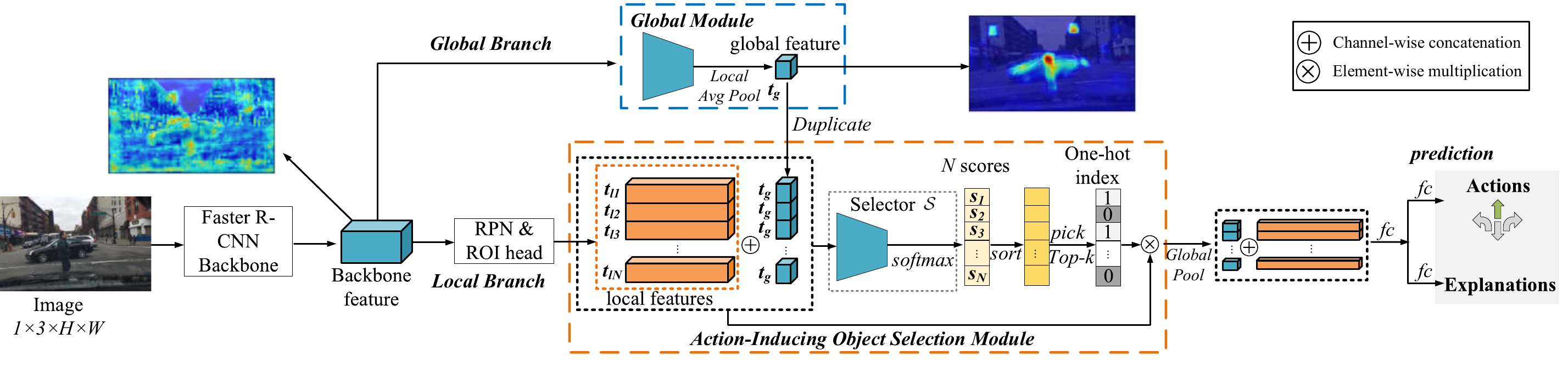}\vspace{-1.0em}
\end{center}
\caption{Architecture of the proposed network.  
The Faster R-CNN is used to extract backbone features, which are fed into a global and a local branch. The Global Module generates a global feature map that provides scene context, while the local branch captures the details of action-inducing objects. In the local branch, a selector module outputs a score for each object feature tensor and associated global context information. The top $k$ action-inducing objects are selected and the features from the two branches are concatenated for action and explanation prediction.
  Two visualizations  derived from the input image are also shown. The combination of local and global features and end-to-end supervision enables the network to reason about scene-object relationships and produce a global feature map more selective of action-inducing objects than the backbone feature maps.}
    \label{fig:net}
    \vspace{-0.8em}
\end{figure*}

\section{Deep Learning Architecture} \label{network}

In this section, we propose a deep architecture for joint prediction
and explanation of object induced actions.

\subsection{Architecture overview}

The prediction of object induced actions and their explanations requires
a combination of several types of reasoning. In this work, we
propose a deep network model based on several steps, which are illustrated in Figure~\ref{fig:net}.
The network initially computes backbone features, which are fed to two modules. The local features $t_{l_i}, {i=1,2,...,N}$ are first produced by the RPN and ROI head layers of Faster R-CNN \cite{ren2015faster}. The Global module generates global features $t_g$ by processing the size and dimension of the backbone features so that they can be combined with the local features as well as modeling scene context and object-scene relationships. Then the local features and global features are fed to Action-Inducing Object Detection module to identify the action-inducing objects. These finally give rise to action
$\hat{A}$ and explanation $\hat{E}$ predictions. The network is
trained with a multi-task loss function 
\begin{equation}
    \mathcal{L} = \mathcal{L}_A + \lambda\mathcal{L}_E,
    \label{eq:loss}
\end{equation}
where $\mathcal{L}_A = \sum_{j=1}^4 L[\hat{A}_j, A_j]$ and
$\mathcal{L}_E = \sum_{j=1}^{21} L[\hat{E}_j, E_j]$, $A_j, E_j$ are
the ground-truth labels for the $j^{th}$ action and explanation, respectively,
$L[.,.]$ is the binary cross entropy loss and $\lambda$ an hyperparameter
that controls the relative importance of action and explanation errors.

This formulation has several benefits. First, the multi-task training allows the explicit optimization of the network for the generation of explanations. This is likely to be more effective than deriving explanations by only a posterior, \eg using only heatmaps that highlight image regions responsible for the prediction~\cite{selvaraju2017grad}. Second, because the generation of explanations is formulated as a
classification problem, the optimization problem is fairly simple,
requiring much less data than the training of natural language systems
based on recurrent networks or generative language models~\cite{kim2018textual}.  Finally, due to the multi-task formulation, actions and explanations
can benefit each other. By receiving explicit supervision for the fact
that a car must slow down {\it because\/} a traffic light is red, the
vision system faces a much simpler learning problem than one who is
only told to slow down. It does not have to
figure out on its own the causal relationship between the light being red and having to slow down.

\subsection{Implementation Details} 

\noindent \textbf{Global Module.} This module generates global features $t_g$ from the Faster R-CNN backbone features. It is composed of two convolutional layers with ReLU activation functions plus a local average pooling operation. It reduces the dimensionality of the backbone features from $2048$ to $256$ and the spatial size of its features maps to $7\times 7$, to enable further joint processing of local and global features.

\begin{table*}[t]
    \centering
    \small
    \begin{tabular}{c|cccc|ccc}
    \hline
      $\lambda$ & F & S & L & R & action mF1 & action F1$_{all}$ &
      explanation F1$_{all}$  \\ \hline
      0 & 0.783 & 0.758 & 0.419 &  0.568 & 0.632 & 0.675 &  -\\ 
      0.01 & 0.819 & 0.760 & 0.504 & 0.605 & 0.672 & 0.696 & 0.329\\
      0.1 & 0.784 & 0.769 & 0.562 & 0.627 & 0.686 & 0.709  & 0.371\\
      1.0 & {\bf 0.829} & {\bf 0.781} & {\bf 0.630} &  {\bf 0.634}
                                & {\bf 0.718} & {\bf 0.734} & {\bf 0.422}\\
    %   100 & 0.777 & 0.744 & 0.407 & 0.449 & 0.594 & 0.641 & 0.392 \\
      $\infty$ & - & - & - & - & - & - & 0.418 \\
      \hline
    \end{tabular}
    \vspace{0.6em}
    \caption{
        Action and explanation prediction performance as a function of
        the importance of each task (determined by $\lambda$) on
        the loss of~(\ref{eq:loss}). Labels denote ``move forward'' (F), ``stop/slow down'' (S),
      ``turn/change lane to the left'' (L), and ``turn/change lane to
      the right'' (R).}
    \label{tab:effectexp}
    \vspace{-1.0em}
\end{table*}

\noindent \textbf{Action-Inducing Object Selection Module.} This module is used to pick action-inducing objects from all object proposals produced by the Faster R-CNN. $N$ local feature tensors $t_{l_i}$ of size $7 \times 7$ are first extracted from the proposal locations and concatenated with the global feature tensor $t_{g}$ to form an object-scene tensor $t_{(l+g)_i}$ per object. 
These tensors are then concatenated into a scene tensor of size $N\times c \times 7 \times 7$ where $c=2048+256$. 
A selector $\mathcal{S}$ then chooses the action-inducing objects from this tensor. $\mathcal{S}$ is implemented with three convolutional layers 
and a softmax layer of $N$ outputs, defining a probability distribution over the $N$ objects. Probabilities are interpreted as action-inducing object scores.
The $k$ objects of largest score are then chosen as action-inducing and the associated object-scene tensors $t_{(l+g)_i}$ passed to the next network stage. 

\noindent{\bf Predictions.} These object-scene tensors are then globally pooled, and vectorized into a feature vector, 
which is fed to three fully connected layers to produce action predictions and explanations.

\noindent{\bf Object-scene relations.} Together, the modules above allow the network to reason about scene-object relationships. The global module provides spatial context for where objects appear in the scene and global scene layout. It can be seen as an attention mechanism that combines the backbone feature maps to produce scene features informative of the location of action-inducing objects. This is illustrated in Figure \ref{fig:net}, where we present an image, the average of the feature maps at the output of the backbone, and the average feature map after global module (the dimension reduced from $2048$ to $256$). While the backbone features have scattered intensity throughout the scene, the global feature maps are highly selective for action-inducing objects. 
This effect is complemented by the selector. Since the latter is learned with supervision from the overall loss function $\mathcal{L}$, it chooses object-scene tensors that improve both the action prediction and explanation accuracy. This provides the global feature map with the supervisory signal needed to highlight the relevant objects. All the remaining proposals receive low score at the selector output and are discarded. This tremendously reduces the clutter created by objects that are unimportant to the action predictions.

\begin{table*}[t]
    \centering
    \small
    \begin{tabular}{c|cccc|c|c||c|c}
    \hline
    models & F & S & L & R & mF1 & F1$_{all}$ & explanation mF1 & explanation F1$_{all}$\\\hline
    only local branch & 0.760 & 0.649 & 0.413 & 0.473 & 0.574 & 0.605 & 0.139 & 0.351\\
    only global branch & 0.820 & 0.777 & 0.499 & 0.621 & 0.679 & 0.704 & 0.206 & 0.419\\
    random selection in Selector & 0.823 & 0.778 & 0.499 & 0.637  & 0.685 & 0.709 & 0.197 & 0.413\\\hline     
    select top-5 & 0.821 & 0.768 & 0.617 & 0.625 & 0.708 & 0.720 & \textbf{0.212} & 0.416\\ 
    select top-10 & \textbf{0.829} & \textbf{0.781} & \textbf{0.630} & \textbf{0.634} & \textbf{0.718} & \textbf{0.734} & 0.208 & \textbf{0.422}  \\ \hline
    % full model (ours) & 0.7336 & 0.7183 & 0.7693 & 0.7904 & 0.4891 & 0.2776\\ \hline
    % local selector \cite{wang2019deep} & \\ \hline
    % attention alignment \cite{kim2018textual}& \\ \hline
    \end{tabular}
    \vspace{0.3em}
    \caption{ Action and explanation prediction performance using global and local features. ``Only local branch'' (``Only global branch'') means that the network ignores global (local) features, ``random Selector'' that object features are chosen randomly, and ``Select top-$k$'' that the selection module chooses the $k$ objects of highest score.}
    \label{tab:globallocal}
    \vspace{-1em}
\end{table*}

\begin{figure*}[t]
\begin{minipage}{0.99\linewidth}
\centering
\includegraphics[scale=0.50,trim={0.5cm 1.8cm 0.2cm 1.5cm},clip]{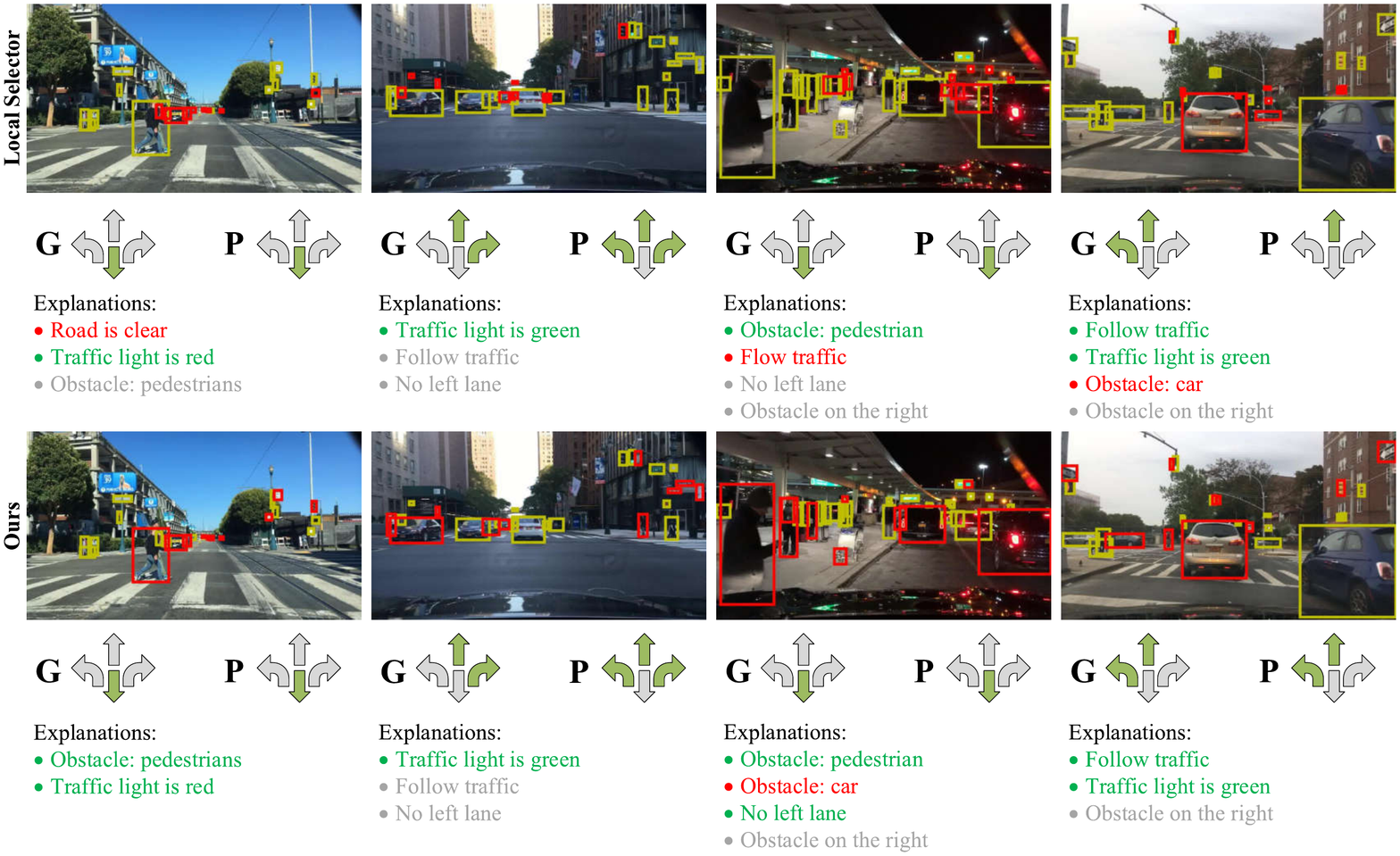}
\caption{Examples of network predictions, objects selected as action-inducing, and explanations. {\color{amber}Yellow bounding boxes} identify the objects detected by the Faster R-CNN, while {\color{red}red bounding boxes} identify the objects selected as action-inducing by the proposed network. "G" stands for ground truth and "P" for prediction. For explanations, {\color{green} green} indicates true positives, {\color{red}red} false positives, and {\color{gray}gray} false negatives (\ie valid explanations not predicted).
}
\label{fig:qualitative}%
\end{minipage}
\vspace{-0.5em}
\end{figure*}

\section{Experiments} \label{experiment}
\subsection{Setup} \label{setup}

All experiments are based on the BDD-OIA dataset. % except Section~\ref{sa_and ma}.
Only the final frame of each video clip is  used, leading to a training set
of 16,082 images, a validation set of 2,270 and a test set of 4,572. The input size of images is $3\times720\times1280$.
The Faster R-CNN is pre-trained on the annotated images from BDD100K \cite{yu2018bdd100k} and frozen
while the remainder of the network of Figure~\ref{fig:net} is trained on BDD-OIA.
The Adam optimizer is used with weight decay of
$1\times10^{-4}$ and initial learning rate $\alpha$ of $0.001$. Training is performed for 50 epochs 
and $\alpha$ divided by 10 every 10 epochs. 
All experiments are evaluated with a standard metric, F1 score, with two variations considered. 
\begin{equation}
    \mathrm{F1}_{all} = \frac{1}{|A|} \sum_{j=1}^{|A|} \mathrm{F1}(\hat{A}_j,A_j), 
\end{equation}
averages the F1 score over all the predictions. Since the dataset is imbalanced, i.e. most of the actions are forward and slow, we further discuss the mean F1 score $m\mathrm{F1}$ of each action $j$ that only compute $\mathrm{F1}(\hat{A}_j,A_j)$ for each sample.

The proposed network is compared to two other models: the
ResNet-101 \cite{he2016deep} (as Baseline) and the network of \cite{wang2019deep}. 
%In \cite{wang2019deep}, they select the top-$k$ only using local object features. Note that, 
Because the latter were designed for tasks other than object induced action recognition,
they are modified to support this task. ResNet-101 is pre-trained on ImageNet. In order to fit ResNet-101 architecture, the input size of images is resized to $3\times224\times224$. Its output layer is modified into 2 branches: a fully-connected (\textit{fc}) layer that outputs 4 action categories, and a \textit{fc} layer that outputs of 21 explanations. The network of \cite{wang2019deep} predicts driving actions. We add to it a new output branch for explanations. All models are trained with the loss of~(\ref{eq:loss}).
The number of action-inducing objects is set to $k = 10$.

\begin{table*}[]
    \centering
    \small
    \begin{tabular}{c|cccc|c|c||c|c}
    \hline
    models & F & S & L & R & mF1 & F1$_{all}$ & explanation mF1 & explanation F1$_{all}$\\\hline
    Baseline &  0.755 & 0.607 & 0.098&0.108& 0.392  & 0.601 & 0.180 & 0.331\\
    % map alignment \cite{kim2018textual} & \\ 
    local selector \cite{wang2019deep} & 0.810 & 0.762 & 0.600 & 0.624 & 0.699 & 0.711 & 0.196 & 0.406\\ \hline
    ours & \textbf{0.829} & \textbf{0.781} & \textbf{0.630} & \textbf{0.634} & \textbf{0.718} & \textbf{0.734} & \textbf{0.208} & \textbf{0.422}  \\ \hline
    % full model (ours) & 0.7336 & 0.7183 & 0.7693 & 0.7904 & 0.4891 & 0.2776\\ \hline
    % local selector \cite{wang2019deep} & \\ \hline
    % attention alignment \cite{kim2018textual}& \\ \hline
    \end{tabular}
    \vspace{0.3em}
    \caption{Comparison of different models.}
    \label{tab:comparetoothers}
    \vspace{-1em}
\end{table*}

\begin{table}[]
    \centering
    % \fontsize{8pt}{8pt}\selectfont
    \scriptsize
    \begin{tabular}{c|cccc|c|c}
    \hline
    Outputs & F & S & L & R & mF1 & F1$_{all}$  \\ \hline
    single action & 0.791 & 0.636 & 0.133 & 0.261 & 0.455 & 0.715\\
    multiple action & 0.795 & 0.680 & 0.522 & 0.594 & 0.648 & 0.665\\ \hline
    \end{tabular}
    \vspace{0.3em}
    \caption{Single \textit{v.s.} multiple action prediction. Single action prediction outputs a single action label given the input image. Multiple action outputs multiple labels.}
    \label{tab:singlevsmulti}
    \vspace{-1.5em}
\end{table}

\subsection{Interplay between Actions and Explanations}

We started by investigating the impact of explanations on action prediction
accuracy. For this, we varied the hyperparameter
$\lambda$ of~(\ref{eq:loss}), as summarized in Table \ref{tab:effectexp}.
Note that $\lambda = 0$ corresponds to ignoring explanations during training
and $\lambda = \infty$ to ignoring action predictions. Interestingly,
the network trained uniquely to predict actions ($\lambda=0$) has the
weakest action prediction performance of all the models. Significant gains
(an increase of action F1$_{all}$ score from 0.675 to 0.734) are achieved
when $\lambda = 1$, \ie when explanations are given as much weight as actions.
This model also has top performance for all action classes.
This shows that explanations are not only useful, {\it but improve the
  performance of autonomous driving system.\/} We believe that this is
the first showing that explainable AI systems can outperform uninterpretable
systems in the vision literature.

Two properties of the proposed explanations justify this observation.
First, the set of explanations is finite and defined based on objects.
This, in turn, enables the robust learning of the explanation
system from a limited set of examples. Open-ended explanation systems, based on
natural language synthesized by recurrent models, lack this property.
Second, and even more critical, the explanations of an object induced
action recognition system are based on causal relations between
objects, \eg ``stop because pedestrians are crossing the street''.
This helps the system learn about object-scene relationships, e.g. figure out
what to localize in the global feature map and relate local to global features, 
enabling a better identification of the action
inducing objects and, consequently, simplifying action predictions.
In the absence of explanations, the system has to figure out all these
relationships by itself.

In summary, for the prediction of object induced actions, the addition of
explanations is manageable and provides direct supervision about the
causality of objects and induced actions that significantly simplify the
learning problem.
This can, in fact, be seen from the results of Table~\ref{tab:effectexp}.
Note that the addition of explanations
produces a much larger gain for actions L and R, the classes of smaller
representation in the dataset (see Table~\ref{tab:all_labels}), than for actions
F and S, the classes or larger representation (a ratio of 2:1 compared to
L and R). This shows that, as the number of training examples
declines and learning has more tendency to overfit, the regularization
due to explanations produces larger gains
in action prediction performance.

\subsection{Interplay between Local and Global Features}

We next tested the importance of combining local and global reasoning.
Table \ref{tab:globallocal} summarizes a series of ablation experiments with
different combinations of local and global features. We started by evaluating
a model that only uses the local features derived
from the Faster R-CNN detections. This achieved the worst performance of
all models tested, for both actions and explanations. The action prediction is highly depended on spatial information without which the accuracy will drop a lot.
We next considered a network using only global features and a network that
 picks the features from $k = 10$ random objects.
While global features performed substantially better than local features, their
performance was slightly weaker than that of the random selection. This suggests
that it is too difficult to predict actions from all the Faster R-CNN object detections. The much improved performance of global features
supports the claim that they enable reasoning about the action-inducing scene parts. 
In fact, global features produced the best explanations
of the three methods. Nevertheless, the slightly better action predictions of random object selection indicate that it is important to consider the objects in detail as well. 
\vspace{-0.1em}

Given all this, it is unsurprising that the combination of the two feature
types resulted in a significant additional performance gain, achieving the
overall best results on the two tasks. This supports the
hypothesis that action prediction requires reasoning about object-scene
interactions. While both the selection of top 5 and top 10 objects, based on
combination of local and global features, outperform all models of a
single feature
type, the number of objects has a non-trivial effect on network
performance. In particular, better results were obtained with 10 than
5 objects. This confirms that BDD-OIA scenes are complex. On the
other hand, the number of objects only had a marginal effect
on explanation performance. In fact, the model with only global
features produced explanations of nearly equivalent quality to those
of the entire network. This suggests that explanations mostly benefit from
contextual reasoning.

\subsection{Model comparisons}

Table \ref{tab:comparetoothers} compares the proposed network to the baseline and the method of \cite{wang2019deep}. The baseline is a purely global method, which predicts actions without extracting object features. It has the worst performance among all methods. This is further evidence for the importance of combining local and global features for action prediction.
The model of \cite{wang2019deep} can be thought of a purely local selector, which uses no global features. Its performance is weaker than the proposed network, and similar to the random selection model of
Table~\ref{tab:globallocal}. Not surprising that this selector lacks the capacity for global reasoning. 
The gains of the proposed network show that object induced action recognition benefits from the analysis of contextual object-scene relationships. 

\subsection{Single vs. Multiple Action Predictions} \label{sa_and ma}
Existing datasests assume that a single action prediction, that chosen by the driver, is possible at each instant. To investigate how this affects action prediction performance, we compared multiple and single action predictions. Single prediction ground truth is computed from IMU data in the original BDD100K dataset, which contains 11,236 training and 3,249 testing images. The network of Figure~\ref{fig:net} is modified to produce a single action prediction, by addition of a softmax layer.  This is compared to the original model, which can predict multiple actions. Table~\ref{tab:singlevsmulti} shows that the performance of each action category is worse for single action predictions. This is for two reasons. First, the IMU labels exacerbate the class imbalance of the dataset. Among training images, there are 6,773 F, 4,258 S, 111 L, and 94 R labels. Seriously imbalanced data lead to models that predict F and S all the time. Second, single labels are deceiving. The fact that the driver chose F does not mean that it was not possible to chose L or R. In result, IMU labels are not ground truth for {\it possible actions,\/} they mostly reflect driver {\it intent\/}. The only conclusion possible from an F label is that the driver {\it wanted\/} to keep moving forward and was not forced to stop, not that F was the only possible action. Again, because a driver typically chooses F and S much more frequently than L or R, the model is encouraged to always predict F or S. In summary, IMU labels encourage autonomous driving vehicles that do not know when it is possible to turn. The introduction of multiple action predictions substantially increases the number of examples per category, mitigating the data imbalance, and creating a lot more examples of scenes with turn labels, mitigating the turn-aversion problem. 

\vspace{-0.2pt}
\subsection{Qualitative Results}

We finally present some qualitative results in Figure~\ref{fig:qualitative}.
In most cases the network predicts actions correctly. One
error is made in the second image, where a left turn is incorrectly predicted as possible. This is likely due to the fact that it is hard to infer left or right turns in the middle of the crossroad.
It is also safe to say that the network can successfully pick the few objects that are action-inducing, including small traffic signs, lights, or obstacles on
the side of the road, while ignoring many other objects
that are not action-inducing.
This is unlike the method of \cite{wang2019deep}, whose selector fails to capture most action-inducing objects, leading to more errors in explanation prediction.

\section{Conclusion} \label{conclusion}
% In this paper, we present our dataset with a new task: to induce possible actions for the vehicle and to pick up appropriate explanations for them. The purpose for the task is to capture most important parts that have influence on driving. Based on our experiments, it is safe to say that we can only count on few objects to make a reasonable driving decision. We prove that, the key to this task is to acquire good contextual information given the specific scenario. We also prove it is necessary to involve both action decision and explanation into this task since those two are closely related and are helpful to improve the performance.
% Original text is above

% {\color{red} 
In this work, we propose the problem of object induced action and explanation prediction for autonomous driving. A dataset was introduced for this task and a new architecture proposed for its solution.
%present our BDD-OIA dataset with a new task: to induce possible actions for the vehicle and to choose appropriate explanations for them. The purpose for the task is to capture most important parts that have influence on driving. 
% Based on our experiments, 
% it is safe to say that we can only count on few objects to make a reasonable driving decision. 
The new task is an interesting challenge for computer vision, because it requires object reasoning that accounts for scene context. The goal is not simply to detect objects, but to detect objects that could create hazard in the autonomous driving setting, and produce explanations for all actions predicted. However, because all explanations are grounded in objects that induce actions, they are easier to generate than in the generic computer vision setting. In fact, they reduce to one-hot style prediction and can be addressed with classification techniques. Due to this, the addition of explanations was shown to increase the accuracy of action predictions in our experiments. We believe that this is the first showing of explanations actually helping improve the performance of a deep learning system.
%is to acquire good contextual information given the specific scenario. It is also necessary to involve both action decision and explanation into this task since they are closely related and are helpful to improve the performance. In conclusion, we propose a new possible paradigm for autonomous driving community.
%For the future work, we plan to involve tracked objects and hierarchical explanations to extend BDD-OIA.
% }

\section*{Acknowledgements}
This work was partially funded by NSF awards IIS-1637941, IIS-1924937, and NVIDIA GPU donations.

{\small
\bibliographystyle{ieee_fullname}
\bibliography{main.bbl}
}

\end{document}